\documentclass{article} 

\usepackage{iclr2025_conference,times}

\usepackage{amsmath,amsfonts,bm}

\def\eqref#1{equation~\ref{#1}}

\def\1{\bm{1}}

\def\ve{{\bm{e}}}

\DeclareMathAlphabet{\mathsfit}{\encodingdefault}{\sfdefault}{m}{sl}
\SetMathAlphabet{\mathsfit}{bold}{\encodingdefault}{\sfdefault}{bx}{n}

\usepackage{hyperref}
\usepackage{url}
\usepackage{colortbl}
\usepackage{booktabs}
\usepackage{multirow}
\usepackage{graphicx}
\title{Are Large Language Models Good In-context Learners for Financial Sentiment Analysis?}

\author{Xinyu Wei$^{1}$, Luojia Liu$^{2}$ \\
$^{1}$Independent Researcher\\
$^{2}$Kenan-Flagler Business School, University of North Carolina at Chapel Hill\\
\texttt{sherryxyw25@gmail.com}, \texttt{josie\_liu@kenan-flagler.unc.edu}
}

\iclrfinalcopy
\begin{document}

\maketitle

\begin{abstract}
Recently, large language models (LLMs) with hundreds of billions of parameters have demonstrated the \textit{emergent ability}, surpassing traditional methods in various domains even without fine-tuning over domain-specific data. However, when it comes to financial sentiment analysis (FSA)---a fundamental task in financial AI---these models often encounter various challenges, such as complex financial terminology, subjective human emotions, and ambiguous inclination expressions. In this paper, we aim to answer the fundamental question: \textit{whether LLMs are good in-context learners for FSA?} Unveiling this question can yield informative insights on whether LLMs can learn to address the challenges by generalizing in-context demonstrations of financial document-sentiment pairs to the sentiment analysis of new documents, given that finetuning these models on finance-specific data is difficult, if not impossible at all. To the best of our knowledge, this is the first paper exploring in-context learning for FSA that covers most modern LLMs (recently released DeepSeek V3 included) and multiple in-context sample selection methods. Comprehensive experiments validate the in-context learning capability of LLMs for FSA.
\end{abstract}

\section{Introduction}
Artificial intelligence has been adopted in various financial applications~\citep{liu2021finrl, nagy2023generative, daluiso2023cva, ganesh2024generative}.
Among these applications, financial sentiment analysis (FSA) is an important task that has attracted substantial attention from both academia and industry~\citep{du2024financial}.
Given online textual sources such as financial reports, earnings calls, and financial news, FSA aims to \textit{assess people’s sentiments and attitudes (e.g., positive, negative, or neutral) toward financial entities, events, or markets}.
Its significance lies in its potential to inform financial forecasting and consequently support business decision-making in the complex and volatile financial market.
In recent years, numerous studies have explored FSA using natural language processing models, particularly with language models that encompass prior knowledge on both finance and sentiment by pretraining on large corpora~\citep{zhang2023instruct, inserte2024large}.

Recently, \textit{large language models (LLMs) with hundreds of billions of parameters} have demonstrated \textbf{emergent abilities}, outshining traditional methods in a variety of domains even without fine-tuning on domain-specific data \citep{zhao2023survey,hwang2025decision,wang2024knowledge}. While LLMs have shown great prowess in sentiment analysis across various domains \citep{althobaiti2022bert, kumarage2024harnessing, ghatora2024sentiment, shaik2024enhancing}, directly applying them to FSA remains challenging. One main reason is the complex financial terminology and expressions. For instance, the words "bull" and "bear" are typically neutral in general vocabulary but can have distinctly positive or negative connotations in financial markets. Additionally, human emotions regarding financial events are inherently subjective. This subjectivity is particularly evident in financial documents, where different individuals may hold varying opinions on the same financial entities or reports. Worse, some people are even reluctant to reveal strong preferences to avoid disrupting the market. Therefore, FSA requires a careful examination of nuanced contexts.

Traditionally, for smaller pre-trained language models (PLMs) with several billions of parameters (e.g., GPT2-1.5B \citep{radford2018improving} and Llama-7B \citep{touvron2023llama}), to overcome the above issues, we can directly fine-tune them via supervised learning on labeled financial text datasets~\citep{zhang2023enhancing, iacovides2024finllama}.
Nevertheless, fine-tuning LLMs for FSA is extremely challenging, if not impossible at all. 
First, a sufficient amount of labeled financial text data is necessary for supervised learning. 
However, since sentiment annotation in finance requires domain-specific knowledge, acquiring enough labeled financial text data may be infeasible in practice.
Second, fine-tuning LLMs is often resource-intensive and time-consuming.
Typically, LLMs are fine-tuned on multiple powerful GPU servers for several days, imposing a heavy burden on resource-limited practitioners. Finally, most modern LLMs with powerful reasoning abilities are closed-source black-box models (e.g., GPT-4o~\citep{achiam2023gpt} and Claude-3.5\footnote{\url{https://www.anthropic.com/claude/sonnet}}), which precludes model fine-tuning in the first place.
Given these challenges, it is natural to ask the following fundamental question: \emph{how can powerful LLMs be adapted for FSA without fine-tuning?}

To answer this question, we explore the potential of in-context learning~\citep{dong2024survey,chen2024fastgas,wang2025mixture} to enhance LLMs for FSA. Specifically, in-context learning enables LLMs to utilize retrieved document-sentiment pairs as in-context demonstrations within the prompt, allowing them to generalize sentiment predictions to new financial documents without requiring model fine-tuning. Through a systematic and comprehensive empirical study, we aim to provide insights into whether modern LLMs can effectively overcome the challenges of FSA using in-context learning. To the best of our knowledge, this is the first comprehensive investigation of the in-context learning capabilities of LLMs with hundreds of billions of parameters for FSA that covers a wide range of modern LLMs—including the Gemini series, GPT series, Llama-405B, Claude 3.5 Sonnet, and the recently released DeepSeek V3—along with multiple in-context sample selection strategies. We conduct extensive experiments across multiple real-world financial datasets using the state-of-the-art LLMs, demonstrating the effectiveness of in-context learning in this domain.

\begin{figure*}[!t]
\setlength {\belowcaptionskip} {-0.2cm}
\centering
\includegraphics[width=\linewidth]{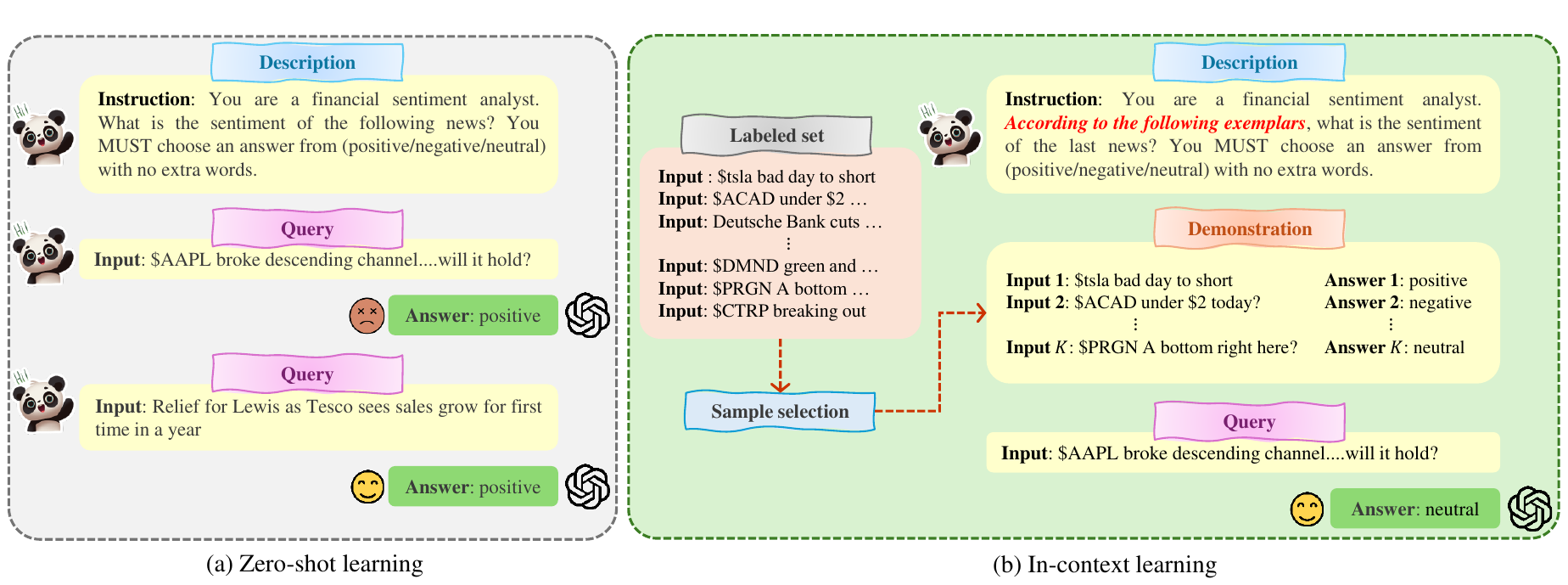}
\caption{Zero-shot learning and in-context learning for financial sentiment analysis.}
\label{fig:framework}
\end{figure*}

\section{Methodology}


\subsection{Problem Formulation}
The focus of this paper is on \textit{in-context learning} for LLM-based FSA, which aims to select financial document-sentiment pairs and include them as demonstrations into LLMs' prompts to support sentiment analysis of new financial documents. Formally, we consider a labeled financial corpora $\mathcal{I} = \left\{\left( q_i, y_i \right) \right\}_{i=1}^{\left| \mathcal{I} \right|}$ where each sample $\left( q_i, y_i \right)$ represents a financial document $q_i$ with its sentiment label $y_i \in \{\texttt{positive}, \texttt{neutral}, \texttt{negative} \}$ annotated by human experts.
The purpose is to select $K$ samples from $\mathcal{I}$ to form a set of demonstrations $\mathcal{I}_{demo}$, where $K$ is a predefined shot of demonstrations.
The selected samples are included in the prompt of LLMs for FSA.

\subsection{Prompt Design}
We design domain-specific prompt templates as the input of LLMs for FSA.
For both zero-shot and in-context learning, the prompt template consists of three parts: (1) background information on the FSA task, (2) specification of the FSA requirement, and (3) restriction of the LLM’s answer to three classes.
Then, we input a specific financial document as a query to LLMs.
Finally, LLMs return one sentiment as the prediction. 
The detailed prompt design is illustrated in Figure~\ref{fig:framework}.

\subsection{In-Context Sample Selection}
In this study, we explore whether LLMs are effective in-context learners for FSA. Given that the key to in-context learning lies in sample retrieval, to comprehensively answer such a research question, we explore the following four representative strategies in our evaluation.

\textbf{\textit{(i)} Random Selection.} 
The most straightforward strategy of sample retrieval is to randomly pick $K$ samples from $\mathcal{I}$.
Generally, random selection is the simplest way to incorporate contextual information for LLMs as in-context learners. However, it can also serve as an \textit{explorer} that provides informative insights into the possible upper bounds of the performance improvement. 

\textbf{\textit{(ii)} Distance-based Selection.} 
The intuition of distance-based selection is to select the most \textit{diverse} samples from $\mathcal{I}$~\citep{hongjin2023selective}, which are expected to cover the most knowledge given that the shot of constraints is usually limited.
To achieve this objective, for each sample $q_i$ in each class, we first employ a PLM to obtain its text embedding $\ve_i$.
Afterward, we select $K$ samples that have the largest distance between each other based on their text embeddings. In the implementation, we first add a random sample to the demonstration set $\mathcal{I}_{demo}$ and then iteratively select the remaining samples from $\mathcal{I}$ that have the largest distances to the text embeddings of samples already in $\mathcal{I}_{demo}$ until $\mathcal{I}_{demo}$ has $K$ samples.
This ensures that the retrieved samples in $\mathcal{I}_{demo}$ are dissimilar from each other, reducing redundancy that may arise from random selection.

\textbf{\textit{(iii)} Difficulty-based Selection.}
The philosophy behind difficulty-based selection is to let LLMs determine which samples are most informative for FSA.
We input all the samples in $\mathcal{I}$ into an LLM and ask it to return the $K$ most difficult ones as the demonstration set $\mathcal{I}_{demo}$, which may be the most informative samples close to the boundary between the three classes.

\textbf{\textit{(iv)} Clustering-based Selection.}
This strategy aims to balance diversity and representativeness in the selected samples.
First, we group the samples from each class in $\mathcal{I}$ into $K$ clusters based on their text embeddings obtained from the PLM.
Afterward, we select the sample closest to the centroid of each cluster.
Consequently, each selected sample represents a specific cluster (i.e., representativeness) while remaining distinct from the others, as they come from different clusters (i.e., diversity).

\begin{table*}[t]
\small
\centering
\caption{Accuracy on \textbf{FiQA} and \textbf{Twitter} datasets. The best-performing method for each LLM is highlighted in \textbf{boldface}.
}
\vskip 0.1in
\label{table:table_main}
\setlength\tabcolsep{12.3pt}
\setlength{\extrarowheight}{1pt}
\begin{tabular}{lccccccccc}
\rowcolor{lightgray!30}\specialrule{1pt}{0pt}{0pt}
\multicolumn{6}{c}{\bf{FiQA}} \\ \specialrule{1pt}{0pt}{0pt}
\bf{LLM}                                & \bf{Zero-shot}    & \bf{Random}   & \bf{Distance}     & \bf{Difficulty}   & \bf{Clustering}   \\ \specialrule{1pt}{0pt}{0pt}
\bf{Gemini 1.5 Flash}                   & 66.00             & 68.67         & 65.33             & 68.67             & \bf{70.67}    \\
\bf{Gemini 2.0 Flash}                   & 66.67             & 69.33         & \bf{72.67}        & 70.00             & 70.00         \\
\bf{Llama 3.1 405B}                     & 71.33             & \bf{74.67}    & \bf{74.67}        & 72.00             & \bf{74.67}    \\
\bf{GPT-3.5 Turbo}                      & 66.67             & 66.67         & 66.00             & 66.00             & \bf{67.33}    \\
\bf{GPT-4}                              & 75.33             & \bf{76.00}    & 74.67             & 73.33             & \bf{76.00}    \\
\bf{GPT-4 Turbo}                        & 68.00             & \bf{76.00}    & 73.33             & 74.67             & \bf{76.00}    \\
\bf{GPT-4o}                             & 69.33             & 68.67         & 72.67             & \bf{76.67}        & 74.67         \\
\bf{GPT-4o mini}                        & 66.00             & 68.00         & 66.00             & \bf{69.33}        & 68.67         \\
\bf{Claude 3.5 Sonnet}                  & 71.33             & \bf{76.00}    & 72.67             & 74.00             & 74.67         \\
\bf{DeepSeek V3}                        & 71.33             & 70.67         & \bf{73.33}        & 67.33             & \bf{73.33}    \\
\rowcolor{lightgray!30}\specialrule{1pt}{0pt}{0pt}
\multicolumn{6}{c}{\bf{Twitter}} \\ \specialrule{1pt}{0pt}{0pt}
\bf{LLM}                                & \bf{Zero-shot}   & \bf{Random}   & \bf{Distance}     & \bf{Difficulty}   & \bf{Clustering}   \\ \specialrule{1pt}{0pt}{0pt}
\bf{Gemini 1.5 Flash}                   & 77.16             & 79.36         & 79.36             & 77.83             & \bf{79.86}    \\
\bf{Gemini 2.0 Flash}                   & 76.65             & 75.80         & 74.45             & 75.80             & \bf{77.33}    \\
\bf{Llama 3.1 405B}                     & 76.82             & 78.68         & 80.03             & 78.68             & \bf{81.73}    \\
\bf{GPT-3.5 Turbo}                      & 75.13             & 76.31         & 76.14             & 76.99             & \bf{79.02}    \\
\bf{GPT-4}                              & 81.22             & \bf{83.25}    & 82.40             & 82.23             & 82.74         \\
\bf{GPT-4 Turbo}                        & 79.36             & \bf{80.37}    & 78.85             & 79.19             & 80.03         \\
\bf{GPT-4o}                             & 83.42             & 81.22         & 82.91             & 83.08             & \bf{84.09}    \\
\bf{GPT-4o mini}                        & 78.34             & 76.99         & 77.50             & 76.48             & \bf{78.85}    \\
\bf{Claude 3.5 Sonnet}                  & 80.37             & 80.03         & 80.54             & 80.37             & \bf{82.74}    \\
\bf{DeepSeek V3}                        & 81.22             & \bf{81.56}    & 79.36             & 79.53             & \bf{81.56}    \\ \specialrule{1pt}{0pt}{0pt}
\end{tabular}
\end{table*}

\section{Experiments}

\subsection{Experimental Setup}
\textbf{Datasets.} 
For the comprehensiveness of our exploration, we adopt two real-world datasets---\texttt{FiQA}\footnote{\url{https://huggingface.co/datasets/pauri32/FiQA-2018}} and \texttt{Twitter}\footnote{\url{https://huggingface.co/datasets/zeroshot/Twitter-financial-news-sentiment}}.
\texttt{FiQA} is a financial dataset adapted from WWW'18 open challenge~\citep{maia201818}. The samples are extracted from financial microblogs, news headlines, and statements. \texttt{FiQA} contains 1,063 training samples and 150 test samples. 
\texttt{Twitter} contains financial news extracted from the tweets using the Twitter API. More information can be found in Appendix~\ref{appendix:datasets}.

\textbf{LLM backbone.}
Moreover, we conduct experiments over ten LLMs with hundreds of billions of parameters from different platforms.
The LLMs that we consider include the Gemini series~\citep{team2023gemini} (i.e., Gemini 1.5 Flash and Gemini 2.0 Flash), Llama 3.1 405B~\citep{dubey2024llama}, the GPT series~\citep{achiam2023gpt} (i.e., GPT-3.5 Turbo, GPT-4, GPT-4 Turbo, GPT-4o, GPT-4o mini), Claude 3.5 Sonnet, and DeepSeek V3~\citep{liu2024deepseek}. For the Llama 3.1-405B and Claude 3.5 Sonnet models, we use the Amazon Bedrock API.
We use \texttt{text-embedding-004}\footnote{\url{https://ai.google.dev/gemini-api/docs/embeddings}} as the PLM to obtain text embeddings.

\textbf{Parameter settings.} In our experiments, we first compare various in-context learning strategies with a default number of shots as 6 with the zero-shot LLMs. We also report the results with varying numbers of shots.
The temperature of LLMs is set to 1.

\subsection{Performance Comparison}
In this part, we compare various in-context learning strategies with the zero-shot LLMs. The results are summarized in Table~\ref{table:table_main}. 
Table~\ref{table:table_main} shows the accuracy of zero-shot learning and in-context learning methods on \texttt{FiQA} and \texttt{Twitter} datasets using different LLMs.
According to the table, we first observe that in-context learning methods almost consistently enhance the LLMs for FSA, which demonstrates that \emph{LLMs are indeed effective in-context learners for FSA}.
In most cases, the zero-shot LLM yields limited performance.
For instance, the zero-shot LLM  has the worst performance by Gemini 2.0 Flash, Llama 3.1 405B, GPT-4 Turbo, GPT-4o mini, and Claude 3.5 Sonnet on \texttt{FiQA}. 
Similarly, the zero-shot LLM performs the worst by Gemini 1.5 Flash, Llama 3.1 405B, GPT-3.5 Turbo, and GPT-4 on the \texttt{Twitter} dataset.
Moreover, we find that the clustering-based selection often outperforms other in-context learning methods, particularly on the \texttt{Twitter} dataset.
We speculate that the improvement achieved by the clustering-based selection strategy stems from its ability to balance diversity and representativeness in the selected samples. 


\begin{figure*}[!t]
\setlength {\belowcaptionskip} {-0.2cm}
\centering
\includegraphics[width=\linewidth]{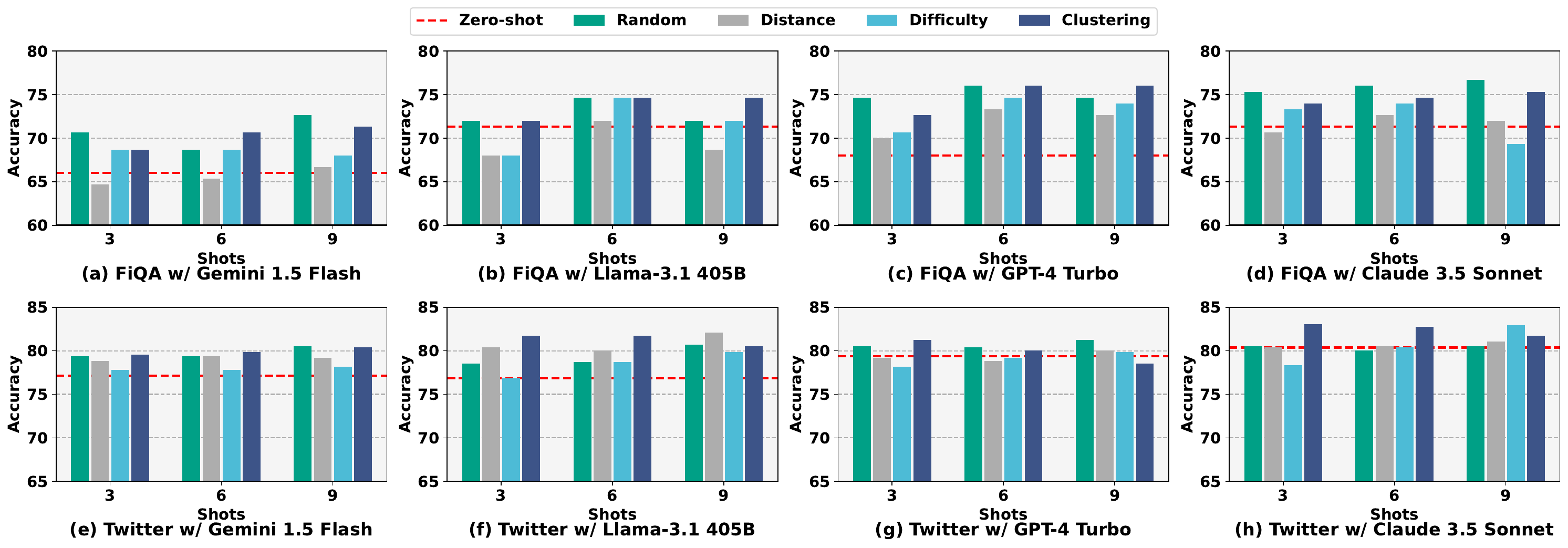}
\caption{Accuracy with varying numbers of shots for both \texttt{FiQA} and \texttt{Twitter} datasets.}
\label{fig:bar}
\end{figure*}

\subsection{Analysis of In-Context Learning Methods}

In this part, we provide deeper insights into the effectiveness of LLMs as in-context learners.

\textbf{Performance with varying numbers of shots.}
We conduct the performance evaluation of LLMs with varying numbers of shots.
Figure~\ref{fig:bar} illustrates the performance of different in-context learning methods with varying numbers of shots.
we observe that the performance of in-context learning does not fluctuate a lot when the number of shots varies.
This suggests that LLMs only require a few demonstrations to learn how to distinguish different sentiments.

\begin{figure*}[!t]
\setlength {\belowcaptionskip} {-0.3cm}
\centering
\includegraphics[width=\linewidth]{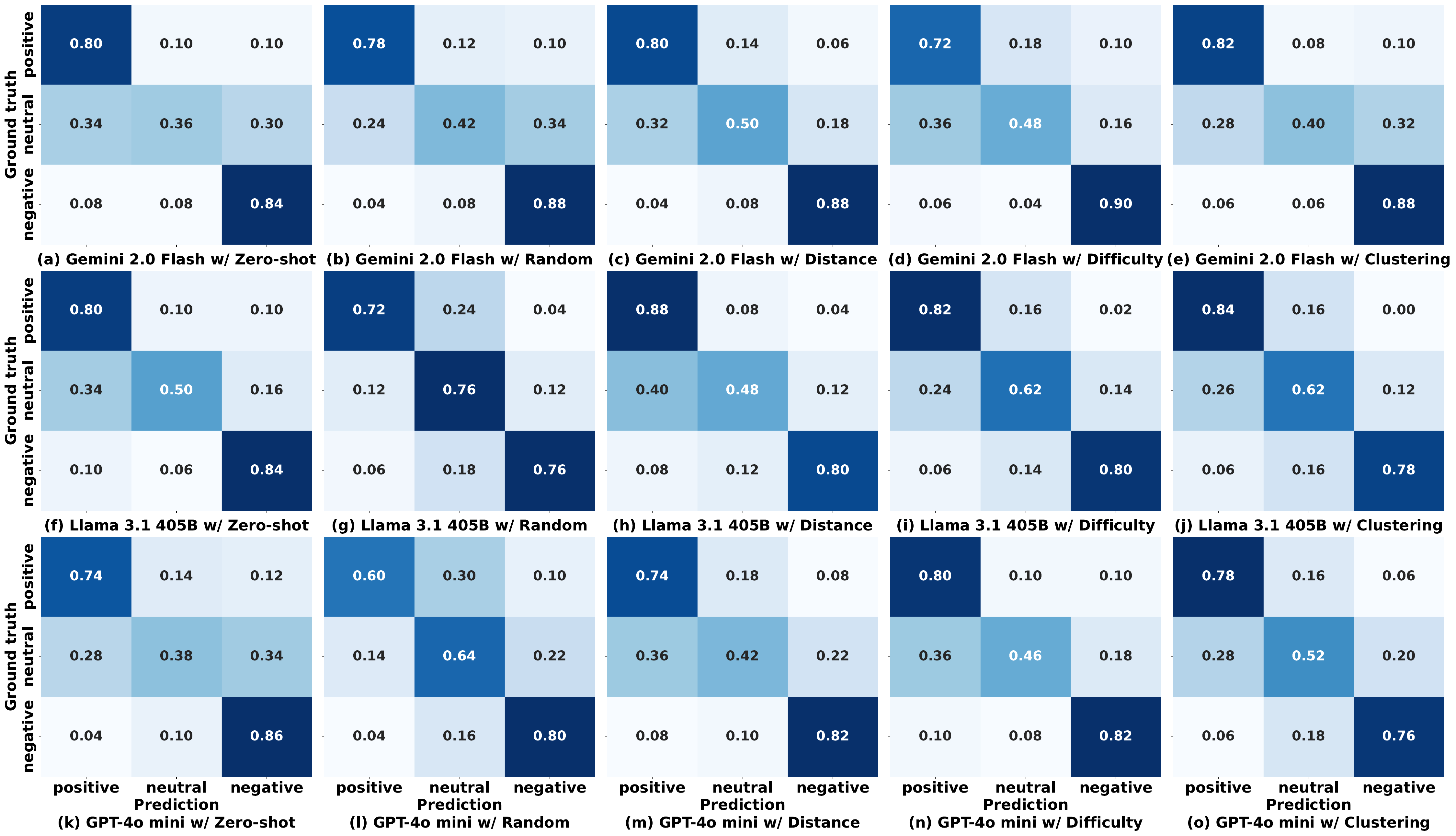}
\caption{Confusion matrices on \texttt{FiQA} with different LLMs and in-context learning methods.}
\label{fig:confusion}
\end{figure*}

\textbf{Performance of neutral samples.}
Figure~\ref{fig:confusion} illustrates the accuracy of zero-shot learning and in-context learning for three classes on \texttt{FiQA}.
According to the figures, we first notice that zero-shot learning can effectively classify positive and negative samples, but neutral samples are usually misclassified.
This suggests that \emph{LLMs struggle to determine the positive/neutral or negative/neutral boundaries}, as the neutral samples are the most ambiguous ones.
However, we can observe that in-context learning methods yield better performance on neutral samples, while the results of positive and negative samples do not diminish.
We conclude that in-context learning provides more information about class boundaries, helping LLMs determine neutral financial samples.

\section{Conclusions}

In this paper, we investigate the fundamental question of whether LLMs with hundreds of billions of parameters are effective \textit{in-context learners} for FSA. Through extensive experiments on two real-world datasets with ten modern LLMs, we demonstrate that in-context learning is a promising solution to address the challenges of the zero-shot LLMs~\citep{yan2023autocast++} and enhance the sentiment prediction accuracy. We also note that the improvement is especially evident when the sentiment associated with the financial document is neutral, which often exhibits the highest ambiguity. Furthermore, our exploration with randomly selected in-context samples highlights the potential for further improvements through more effective in-context sample retrieval strategies.

\bibliography{iclr2025_conference}
\bibliographystyle{iclr2025_conference}

\appendix

\section{Datasets} \label{appendix:datasets}
We use two real-world financial datasets to evaluate the performance of in-context learning for FSA.
The statistics of these datasets can be found in Table~\ref{table:table_datasets}.

\begin{table*}[ht]
\small
\centering
\caption{The statistics of FiQA and Twitter.
}
\vskip 0.1in
\label{table:table_datasets}
\setlength\tabcolsep{12.4pt}
\setlength{\extrarowheight}{3pt}
\begin{tabular}{crrrr}
\rowcolor{lightgray!30}\specialrule{1pt}{0pt}{0pt}
\multicolumn{5}{c}{\bf{FiQA}} \\ \specialrule{1pt}{0pt}{0pt}
                & \bf{Positive}     & \bf{Neutral}      & \bf{Negative}     & \bf{Total}    \\ \specialrule{1pt}{0pt}{0pt}
\bf{Train}      & 632               & 34                & 295               & 961           \\
\bf{Test}       & 50                & 50                & 50                & 150           \\
\bf{Total}      & 682               & 84                & 345               & 1,111         \\
\rowcolor{lightgray!30}\specialrule{1pt}{0pt}{0pt}
\multicolumn{5}{c}{\bf{Twitter}} \\ \specialrule{1pt}{0pt}{0pt}
                & \bf{Positive}     & \bf{Neutral}      & \bf{Negative}     & \bf{Total}    \\ \specialrule{1pt}{0pt}{0pt}
\bf{Train}      & 1,923             & 6,178             & 1,442             & 9,543         \\
\bf{Test}       & 475               & 1,566             & 347               & 2,388         \\
\bf{Total}      & 2,398             & 7,744             & 1,789             & 1,1931         \\ \specialrule{1pt}{0pt}{0pt}
\end{tabular}
\end{table*}

\section{Related Work}
\subsection{Financial Sentiment Analysis}
FSA aims to evaluate humans' attitudes (e.g., positive, negative, or neutral) toward financial entities, events, or markets. Early FSA studies relied on lexicon-based approaches~\citep{oliveira2016stock} to identify sentiment-bearing words such as Loughran-McDonald Sentiment Word Lists~\citep{loughran2011liability}. However, they struggle with automation, scalability, and interpretation accuracy. As computational techniques advanced, machine learning-based approaches, including random forests~\citep{talazadeh2024sarf}, Naive Bayes~\citep{maragoudakis2016exploiting}, and support vector machines (SVMs)~\citep{le2015Twitter, onwuegbuche2019support}, became widely adopted for the classification of sentiment indicators in financial reports, news articles, and earnings calls. More recently, deep learning models, such as long short-term memory (LSTM)~\citep{jangid2018aspect, jiawei2019stock, li2019dp, swathi2022optimal}, convolutional neural networks (CNNs)~\citep{sohangir2018big} and graph neural networks (GNNs)~\citep{huang2022graph}, are employed to improve classification accuracy in financial language. 
However, given the complexity of contextual sensitivity, finance-specific terminology, and the high volume of market data, traditional sentiment analysis lacks the sophistication to capture nuanced market dynamics and contextual shifts. 
Motivated by this, an increasing number of research initiatives are navigating toward LLMs to enhance financial text analysis.

\subsection{Language Models for Financial Sentiment Analysis}
Recently, substantial efforts have been devoted to applying PLMs to financial sentiment analysis.
For instance, FinBERT~\citep{liu2021finbert} pre-trains a BERT model~\citep{kenton2019bert} over financial data and fine-tunes it for diverse financial downstream tasks, including FSA.
\citet{zhang2023enhancing} enhance FSA via retrieval augmented PLM.
They first apply instruction tuning~\citep{wei2022finetuned} to fine-tune an LLM to align its behavior with predicting financial sentiment labels and subsequently generate a sentiment prediction based on retrieved information from external sources.
FinLLAMA~\citep{iacovides2024finllama} fine-tunes Llama-7B~\citep{touvron2023llama} on a combination of four labeled publicly available financial datasets to benefit algorithmic trading.
While the aforementioned methods achieve strong performance in FSA, they require a large amount of labeled financial data and powerful GPU servers for fine-tuning PLM.
In contrast, we focus on modern LLMs with hundreds of billions of parameters and attempt to investigate whether they are good in-context learners for FSA.


\subsection{In-Context Learning}
In-context learning~\citep{dong2024survey} enables LLMs to quickly adapt to new tasks for predicting the target of a given query without fine-tuning by providing a few query-target pairs (i.e., demonstrations).
Despite this incredible capability, the effectiveness of in-context learning is highly sensitive to the selection of demonstrations~\citep{zhao2021calibrate}.
A naive strategy is to randomly select samples as the demonstration set. Although random selection is straightforward, it does not ensure the selected samples are informative and helpful for LLMs.
Previous studies have highlighted the challenges and instability of demonstration selection.
For example, one effective method is to select the most diverse samples by choosing samples whose text embeddings are most distant from each other~\citep{hongjin2023selective}.
Another approach is to let LLMs decide which samples should be selected.
Although a few studies investigate the performance of in-context learning for different PLMs~\citep{dmonte2024evaluation}, a thorough evaluation of LLM performance in in-context learning for FSA is still missing.


\end{document}